\documentclass[twocolumn, a4paper]{article}
\usepackage[margin=25mm]{geometry}
\usepackage{subcaption}
\usepackage{booktabs}
\usepackage{multicol}
\usepackage{graphicx}
\providecommand{\keywords}[1]
{
  \small	
  \textbf{\textit{Keywords---}} #1
}

\date{}

\begin{document}

\title{Oscillatory Neural Network as Hetero-Associative Memory for Image Edge Detection}

\author{Madeleine Abernot \\
    LIRMM, Univ. of Montpellier,\\ CNRS \\
    \texttt{madeleine.abernot@lirmm.fr}
    \and
    Thierry Gil \\
    LIRMM, Univ. of Montpellier,\\ CNRS \\
    \texttt{thierry.gil@lirmm.fr}
    \and
    Aida Todri-Sanial \\
    LIRMM, Univ. of Montpellier,\\ CNRS \\
    \texttt{aida.todri@lirmm.fr}
}

\maketitle

\begin{abstract}
The increasing amount of data to be processed on edge devices, such as cameras, has motivated Artificial Intelligence (AI) integration at the edge. 
Typical image processing methods performed at the edge, such as feature extraction or edge detection, use convolutional filters that are energy, computation, and memory hungry algorithms. 
But edge devices and cameras have scarce computational resources, bandwidth, and power and are limited due to privacy constraints to send data over to the cloud. 
Thus, there is a need to process image data at the edge.  Over the years, this need has incited a lot of interest in implementing neuromorphic computing at the edge. 
Neuromorphic systems aim to emulate the biological neural functions to achieve energy-efficient computing.

Recently, Oscillatory Neural Networks (ONN) present a novel brain-inspired computing approach by emulating brain oscillations to perform auto-associative memory types of applications. 
To speed up image edge detection and reduce its power consumption, we perform an in-depth investigation with ONNs. 
We propose a novel image processing method by using ONNs as a hetero-associative memory (HAM) for image edge detection. 
We simulate our ONN-HAM solution using first, a Matlab emulator, and then a fully digital ONN design. We show results on gray scale square evaluation maps, also on black and white and gray scale 28x28 MNIST images and finally on black and white 512x512 standard test images. 
We compare our solution with standard edge detection filters such as Sobel and Canny. 
Finally, using the fully digital design simulation results, we report on timing and resource characteristics, and evaluate its feasibility for real-time image processing applications.
Our digital ONN-HAM solution can process images with up to 120x120 pixels (166 MHz system frequency) respecting real-time camera constraints.
This work is the first to explore ONNs as hetero-associative memory for image processing applications.
\end{abstract}



\section{Introduction}

In the last decade, we have witnessed a fast proliferation of edge devices for personal use and in all industry sectors. In particular, smart edge cameras are widely used for multiple applications such as security, automotive and human-computer interaction among others \cite{edgecomputing}. They perform image capture and processing to identify an image or detect objects \cite{SmartCameras}. For privacy reasons, edge devices recently have some sort of Artificial Intelligence (AI) embedded by using Artificial Neural Networks (ANNs). But the increase of data to be processed combined with limited computational resources, bandwidth, and energy at the edge, has led to exploring novel beyond-von Neuman computing paradigms inspired by the brain, such as neuromorphic computing.

Oscillatory Neural Networks (ONNs) \cite{ONN1, ONN2, ONN3, ONN4} are a novel neuromorphic computing paradigm based on coupled oscillators to mimic brain waves observable on electroencephalogram (EEG) \cite{eeg}. Information is represented in the phase relation between oscillators to limit voltage amplitude and allow low-power computation \cite{lowpower}. Coupled oscillators exhibit rich dynamics for Auto-Associative Memory (AAM) tasks \cite{onn_pat}, like in Hopfield Neural Networks (HNNs) \cite{hopfield}. Similarly, as HNNs, thanks to their AAM feature, ONNs are commonly used for pattern recognition applications \cite {OnnFPGA}. But, ONNs can be devised to perform other functionalities beyond AAM. For example, \cite{legion} proposed a solution for image segmentation using a network of oscillators. Another work, \cite{optimization} developed a large-scale ONN to resolve combinatorial optimization problems. In this work, we propose a new functionality on ONNs such as Hetero Associative Memory (HAM), which can be advantageous for image edge detection applications.

Edge detection is an image processing function detecting brightness and color variations in images. It also helps for more complex image processing operations like feature extraction, image classification, image segmentation, or object detection. Usual algorithms are based on convolutional filters, such as in Convolutional Neural Networks (CNNs) \cite{cnn}.

CNNs are state-of-the-art ANNs used for image processing applications. They are multi-layer ANNs with first layers comprising convolutional filters to extract features from images, like edges. Previous work from \cite{corti} proposes a solution to use ONN as a CNN first-layer filter. They use ONN as AAM to perform edge detection on 28x28 MNIST images.

Here, we propose for the first time a new solution to perform edge detection using ONN as HAM. It reduces the number of parameters compared to convolutional edge detection filters. 
Moreover, this work opens up ONNs to other image processing applications using multi-dimensional association. Our contributions can be summarized as 1) development of a solution to use ONN as HAM for image edge detection application, 2) the development and simulation of our solution using an HNN-based Matlab emulator on multi-scale black and white and gray scale images, and 3) a study of the real-time performances of the hetero associative ONN on edge detection, using a fully-digital ONN design from \cite{OnnFPGA} simulated on FPGA.

The paper is organized as follows. In Section~\ref{sec:ONN} we describe the ONN paradigm and its AAM computation capabilities. In Section~\ref{sec:HAM} we discuss the ONN adaptation to perform HAM tasks. Next, in Section~\ref{sec:edgeDetect} we present our hetero associative ONN solution to perform edge detection. The Section~\ref{sec:designs} describes the Matlab emulator and the digital design used for simulation. Then, in Section~\ref{sec:Results} we show results obtained with the HNN-based Matlab emulator and the digital ONN design on multiple black and white and gray scale images. Additionally, we compare them with standard Sobel and Canny edge detection algorithms. 
Finally, in Section~\ref{sec:Discussion} we provide timing and resource characteristics and discuss advantages and limitations of our ONN-HAM solution. 

\section{Oscillatory Neural Networks}
\label{sec:ONN}

ONNs are a novel brain-inspired computing paradigm based on coupled oscillators. 
This section details the ONN computing paradigm, its ability to deal with AAM tasks, and the associated learning algorithm.  

\subsection{Computing Paradigm}
\label{sec:ONNcompPar}

\begin{figure}
    \centering
    \includegraphics[width=.95\linewidth]{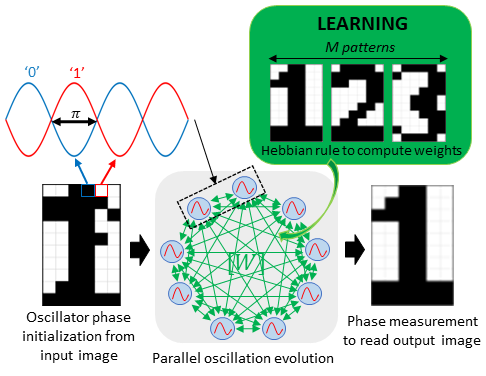}
    \caption{An oscillatory neural network representation with AAM type of computation.}
    \label{fig:AAM-ONN}
\end{figure}

\begin{figure}
    \centering
    \includegraphics[width=.95\linewidth]{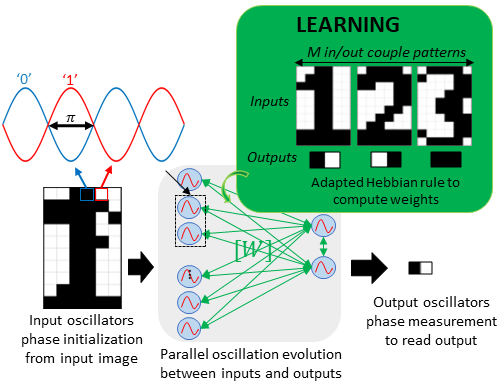}
    \caption{An oscillatory neural network representation with HAM type of computation.}
    \label{fig:HAM-ONN}
\end{figure}

ONNs are recurrent neural networks made of coupled oscillators where each neuron is an oscillator coupled with analog components such as resistors or capacitors. 
The phase relationship between oscillators encodes the information.
For example, for binary information, an oscillator with a $0^o$ phase difference from a reference oscillator represents a logic '0', and an oscillator with a $180^o$ phase difference represents a logic '1', see Fig.~\ref{fig:AAM-ONN}. 

ONN exploits the switching dynamics between oscillators \cite{onn_phys} to compute.
The process starts by defining the coupling between oscillators with the learning algorithm, which depends on the task to be executed. Afterward, the input information defines oscillators' phases to initialize the network. The network evolves until it stabilizes, and once it is stable, the output information is obtained by measuring and decoding the oscillators' phases.

ONNs appear as good solutions for hardware implementation on edge devices due to the fast and energy-saving computation \cite{lowpower}. The phase computation allows to use signals with low voltage amplitude to limit the power consumption. As well, the highly parallel dynamic computation of oscillators enables fast computation, proportional to the oscillators' frequency.

\subsection{Auto-Associative Memory (AAM)}
\label{sec:ONNAAM}

The coupling between oscillators defines the ONN behavior. It has been shown that using a specific coupling, ONNs can compute AAM tasks such as pattern recognition applications \cite{onn_pat}, like with HNNs \cite{hopfield}. An AAM network can memorize patterns in a way that it can retrieve one of the memorized patterns from corrupted input information. 
For example, ONN initialized with a corrupted pattern will evolve for various oscillating cycles and stabilize in one of the memorized patterns, see Fig.~\ref{fig:AAM-ONN}. In an image, each oscillator corresponds to a pixel, whereas the oscillator's phase represents the pixel's color. 
If we initialize the network with any image (corrupted or not), it will evolve and stabilize to the closest memorized image.
Note, in AAM, input and output have the same dimension and are represented by the same neurons. 
Also note, it is possible for the network to never stabilize to a memorized pattern. In this case, we consider the input image as unstable. 
The main advantage of ONNs over HNNs is the analog implementation allowing parallel low-power computing \cite{lowpower}.

\subsection{Learning}
\label{sec:ONNAAMLearn}

The predominant learning algorithm used to perform AAM tasks is the Hebbian learning rule \cite{hebbian}. It is an unsupervised learning rule introduced by Hopfield to resolve AAM tasks on HNNs. For a network of $N$ neurons, to learn $k$ patterns $X^k$. The unsupervised Hebbian learning rule defines the weight $w_{ij}$ between neuron $i$ and neuron $j$ as:

\begin{equation}
    w_{ij} = \frac{1}{k}\sum_{k}{X_{i}^{k} {X_{j}^{k}}^T}
\end{equation}

with $w_{ij} = 0$ $\forall$ $i = j$. Note that the Hebbian learning rule was developed for learning binary (0,1) or bipolar (-1,1) patterns. Thus, for image processing, it can learn only black and white images. However, ONN uses oscillator's phases to encode information, so it can tolerate learning and stabilization of patterns with additional values corresponding to phases among $0^o$ and $180^o$.
Also note, the Hebbian learning rule enables ONN to retrieve memorized patterns and opposite ones.

\section{Hetero-Associative Memory (HAM)}
\label{sec:HAM}

HAM are networks executing associative memory tasks between inputs and outputs with different dimensions.
Inputs and outputs use different neurons to represent different information. HAM can be considered as a two-layer network that associates input information with output information, or reconstructs in-out pairs, see Fig.~\ref{fig:HAM-ONN}. For example, it can associate an input image with an output class. Thus, learning algorithms use patterns of in-out pairs.

In this section, we first discuss existing HAM neural networks and their limitations. Then, we explain how to modify ONNs to allow HAM functionality. Finally, we present the learning algorithm adapted for hetero association with ONNs.

\subsection{State-of-the-Art}
\label{sec:HAMSotA}

The principle of the HAM network was presented by Kohonen in \cite{ham1} and is referred to as the Linear Associative Memory (LAM). It uses perceptron neurons to process sequentially as a two-layer feed-forward network with input influencing output depending on the synapse's weights. It is a "one-shot" memory association with limited accuracy. To overcome this limitation, Kosko introduced the Bidirectional Associative Memory (BAM) \cite{kosko}. It sequentially uses both forward and backward communications between input and output layers to perform hetero association. Using a simple adapted Hebbian learning algorithm, they were able to perform HAM tasks better than LAM.
Intrinsically, ONNs compute in parallel and do not support feed-forward computation due to always coupled oscillators. Such as if neurons from input and output layers are coupled, they will interact and evolve in parallel. This parallel computation takes advantage of both forward and backward communications for fast hetero association computation. Therefore, we need to adapt the standard ONNs to enable hetero association functionality.

\subsection{ONN Adaptation}
\label{sec:ONNHAM}

The primary constraint to performing HAM tasks with ONN is the coupling among all oscillators. In both LAM or BAM, only neurons from the input layer are initialized, and the network figures out the output. But in ONNs, all coupled oscillators interact simultaneously. Thus, if only neurons of the input layer are initialized, neurons of the output layer will also impact the input layer due to coupling. 
A solution to counter this effect is to initialize output oscillators with neutral values, such that they do not influence the dynamic. 
Presently, we only use bipolar output values with our ONN (-1,+1) due to the Hebbian learning algorithm limitation. Consequently, to avoid the influence of one of the bipolar values, we use two neurons for each output information and initialize one with -1 and the other with +1. Hence, one output has the same influence for each possible bipolar value. Note, we interconnect output neurons as they represent the same information. Also note, the learning algorithm will use patterns with doubled output information.

\subsection{Learning}
\label{sec:ONNHAMLearn}

For HAM applications, learning is performed with multi dimensional in-out data pairs such as image-class pairs. The learning algorithm to train HAM is an adapted Hebbian learning rule. It reproduces the Hebbian algorithm while connections among input and output neurons are removed. If we consider $k$ pairs of patterns with $i$ neurons in the input layer $X$ and $j$ neurons in the output layer $Y$, the weight $w_{ij}$ between input neuron $i$ and output neuron $j$ is: 

\begin{equation}
    w_{ij} = \sum_{k}{X_{i}^{k}Y_{j}^{k}}
\end{equation}

To keep the connections between output neurons, we translate each in-out pair pattern into a unique vector. Then, we compute the classical Hebbian rule of Sec.~\ref{sec:ONNAAMLearn}, but we set to $0$ all connections among the input neurons.

\section{Edge Detection with ONN}
\label{sec:edgeDetect}

Edge detection extracts edges from image contrasts. It typically occurs between two regions of an image. It is often employed for image pre-processing like image filtering, or feature extraction (like corners, curves, etc.). This section discusses existing edge detection algorithms, and we present the ONN adaptation to filter the image with the chosen learning patterns. 

\subsection{State-of-the-art}
\label{sec:SoTA}

There exist different types of edge detection algorithms. 
Commonly used edge detection algorithms \cite{edge}, such as Sobel \cite{sobel} and Canny \cite{canny} algorithms, use convolutional kernels to process. Kernels are small-size matrices (usually from 3x3 to 7x7) whose parameters are applied as convolutional operators on a small part of the image. Convolutional results are then used to calculate a gradient of the small image part which identify if there is an edge. Another step applies a threshold to the obtained gradient to select only strong edges. Each kernel scans the entire image to identify edges at each pixel location of the image. So, these algorithms are scalable to every image size. 

Both Sobel and Canny algorithms use at least two kernel matrices to detect edges, one to detect horizontal edges and the other for vertical edges. Note that other solutions exist with more than two filter matrices to increase precision. Also note, Canny algorithm adds a pre-processing gaussian filter to remove noise and smooth images. 

\subsection{ONN adaptation}
\label{sec:ONNAdaptEdge}

\begin{figure}
    \centering
    \includegraphics[width=0.95\linewidth]{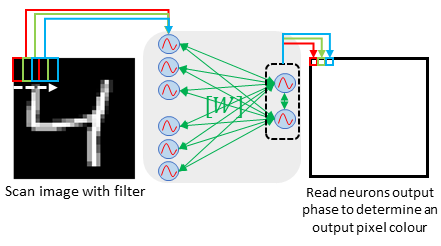}
    \caption{Complete image filtering by scanning with ONN-HAM solution.}
    \label{fig:ONN-scan}
\end{figure}

We adapt ONN to perform image edge detection by using it as a 3x3 filter with a 1-pixel stride, see Fig.~\ref{fig:ONN-scan}. Note, the image size is scaled down as we scan without any padding. 
We define the ONN input with a 3x3 image and the ONN output with the detected edge. We configure the network to detect two types of edges represented by 1 bipolar output information, such as -1 represents one type of edge, and +1 represents another type. However, to deal with the oscillators' parallel dynamics as explained in Section~\ref{sec:ONNHAM}, we use two neurons at the output layer to represent each bipolar output.

\begin{figure}
    \centering
    \includegraphics[width=0.85\linewidth]{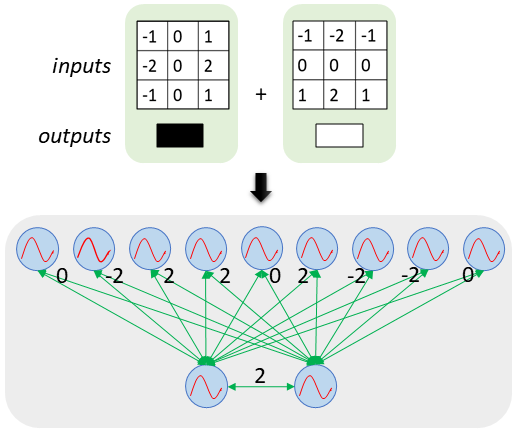}
    \caption{ONN-HAM training patterns and weights for image edge detection application.}
    \label{fig:ONN-TP}
\end{figure}

Once the ONN architecture is defined, it remains to choose training patterns. We first explored multiple combinations of bipolar vertical, diagonal, and horizontal edges to find the best training patterns. However, we finally found out the two kernels used for Sobel convolution can efficiently be used as training patterns to define weights, see Fig.~\ref{fig:ONN-TP}. One pattern associates the horizontal edge kernel with white pixels output (+1, +1), and the second associates the vertical edge kernel with black pixels output (-1, -1). We also expect ONN to detect a no-edge case by stabilizing to a non-memorized output, like (+1, -1) or (-1, +1).

\section{ONN-HAM designs}
\label{sec:designs}

We validate and test our ONN-HAM solution with two ONN designs adapted for hetero-association. A first ONN Matlab emulator design based on HNN, described in Section~\ref{sec:MatEmulator}, and a second fully digital ONN design specified in Section~\ref{sec:DigDesign}.

\subsection{ONN Matlab emulator}
\label{sec:MatEmulator}

We first test our ONN-HAM solution with a Matlab emulator based on HNN.
HNN state evolution is defined by the following update rule.
For each neuron $i$ with active state $s_i^t$, connected to $j$ neurons with synaptic weights $w_{ij}$, the next neuron state is:
\begin{equation}
    s_{i}^{t+1} = sign(\sum_{j}{w_{ij}s_{i}^{t})}
\end{equation}

The Matlab emulator uses a fully connected 11-neuron HNN where the 9 first neurons represents the 3x3 input image, and the 2 last neurons represents the output, using synaptic weights from Fig.~\ref{fig:ONN-TP}.
Note, this network only outputs binary or bipolar values. However, neuron initialization can be done with in-between values, as necessary for gray scale images.

\subsection{ONN fully digital design}
\label{sec:DigDesign}

After validation of the ONN-HAM solution with the Matlab emulator, we simulate it with a fully digital ONN design.
The design was presented in \cite{OnnFPGA}. It took inspiration from a hybrid analog-digital design proposed by \cite{jackson}. 
The fully digital design uses phase-controlled digital oscillators to emulate neurons and a memory table to emulate synapses weights.
In this design, each oscillators period is encoded with 16 clock cycle period, allowing 9-stage phases between $0^o$ and $180^o$ (from 0 to 8). So, each neuron (image pixel) can be initialized with up to 9 different gray levels.
To do so, gray scale images are converted into 9 gray levels encoded into the different oscillator's phases.
Note the oscillators' frequency is defined as $F_{osc} = F_{sys} / (16*4)$ with $F_{sys}$, the system frequency, to ensure system computation.

Like with the Matlab emulator, we use an 11-neuron fully connected ONN with 9 neurons as input and 2 neurons as output, using synaptic weights from Fig.~\ref{fig:ONN-TP}.
We use the Vivado design tool to simulate the digital ONN-HAM design with the XC7Z020-1CLG400C FPGA as target device.

\section{Results}
\label{sec:Results}

To test our ONN-HAM solution, we perform edge detection first on gray scale square evaluation maps, then on black and white and gray scale 28x28 MNIST images, and finally on large scale 512x512 black and white standard images. We compare outputs obtained with our two ONN-HAM designs with the standard Sobel and Canny algorithms.
In this section, we present results obtained on the different images while performing edge detection with our ONN-HAM solution using both the Matlab emulator and the fully digital design.

\subsection{Gray scale maps}
\label{sec:GrayMaps}

Quality evaluation of edge detection algorithms is complex because a ground truth is necessary to compare with. 
However, up to our knowledge, there is no approved algorithm to define the ground truth. 
Thus, image edge detection ground truth is rarely available.

\begin{figure}
    \centering
    \includegraphics[width=.90\linewidth]{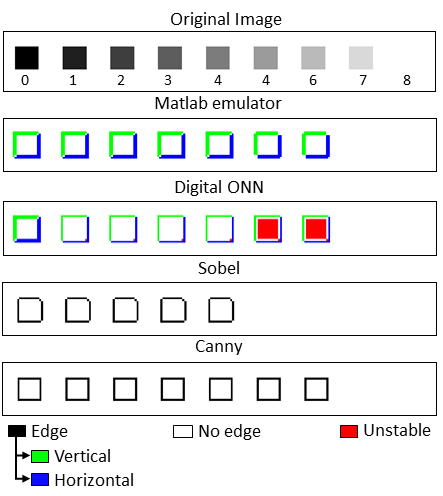}
    \caption{Comparison of number of output images obtained with different edge detection solutions on an square gray scale evaluation map}
    \label{fig:sub1}
\end{figure}

Consequently, to evaluate edge detection algorithms, \cite{EdgeMaps} proposed to use evaluation maps representing simple forms on a fixed background with different levels of gray and different levels of noise. 
As a first evaluation method of our solution, we create an evaluation map with white background, and gray square forms with 9 levels of gray, from 0-black square, to 8-white square, see Fig.~\ref{fig:sub1}.
Note, we do not provide noise resistance analyses in this work.

Fig.~\ref{fig:sub1} shows output images given by the two ONN-HAM designs and two state-of-the-art edge detection algorithms, Sobel and Canny.
It is important to highlight that the ONN-HAM solution correctly identifies both vertical and horizontal edge categories, as well as the no-edge category with the two training patterns. ONN sometimes retrieves horizontal edges when vertical and the other; however, this is not an issue for a global edge detection process.

\begin{figure}
    \centering
    \includegraphics[width=.99\linewidth]{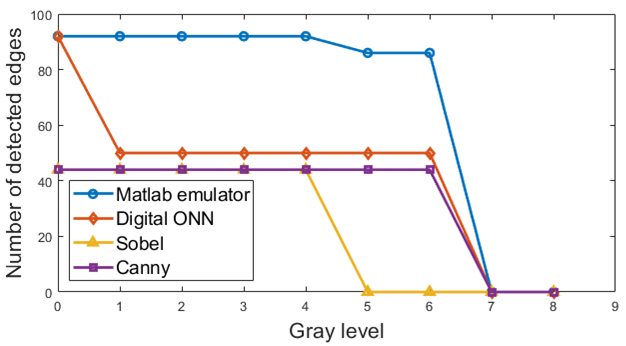}
    \caption{Comparison of detected edges obtained with different edge detection solutions on an square gray scale evaluation map.}
    \label{fig:sub2}
\end{figure}

More precisely, on the 0-black level, edges are correctly retrieved. However, Fig~\ref{fig:sub2} shows our solution detects twice each detected edge. 
Note the Sobel and Canny algorithms avoid this double detection thanks to the threshold function which select only strong edges.
Then, on gray scale levels, the ONN-HAM solution can detect edges until the $6^{th}$ level of gray, as the Canny algorithm, while the Sobel algorithm only detects edges until the $4^{th}$ level of gray.
From this point of view, the ONN-HAM solution presents higher efficiency on gray scale edges than Sobel algorithm.
Note, the two ONN-HAM designs give equal results for black and white edges, while the fully digital design gives less edge than the Matlab emulator for gray scale edges.
Also note, some gray scale edges are unstable with the fully digital design. 
It means, for some gray scale images, the ONN-HAM never stabilizes to a stable output, and hesitates between multiple patterns. 
We consider unstable outputs as no-edge category as the network does not stabilize to one of the training pattern.

\subsection{28x28 MNIST Images}
\label{sec:Res28x28Mat}

\begin{figure}
    \includegraphics[width=0.8\linewidth]{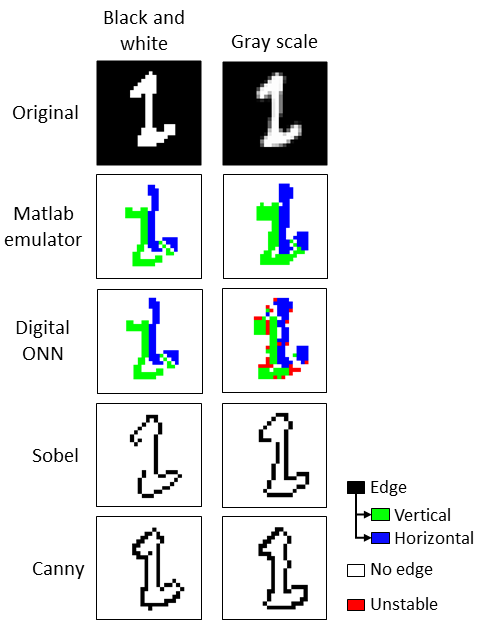}
    \caption{Output images obtained with different edge detection solutions on a 28x28 black and white and gray scale MNIST image.}
    \label{fig:ONN-MNIST-Matlab}
\end{figure}

After a first evaluation on a gray scale evaluation map with only square forms, we use our ONN-HAM solution on more complex and realistic images from MNIST database \cite{mnist}.
The MNIST database contains 28x28 gray scale images representing handwritten digits. 
We perform edge detection tests on both gray scale and binarized black and white images. Fig.~\ref{fig:ONN-MNIST-Matlab} shows our ONN-HAM solution (Matlab emulator or fully digital design) which correctly identifies most digits edges on the black and white image. Using only two training patterns representing vertical and horizontal edges, the network also identifies diagonal edges. Only left-oriented diagonal is not retrieved every time.
Like with the evaluation map, most detected edges are detected twice. 
On the contrary, Sobel algorithm misses some horizontal edges.
Again, our ONN-HAM solution seems more efficient to detect all edges than the Sobel algorithm.

However, for gray scale images, the double detection phenomena and the high efficiency to correctly detect gray scale edges (as identified in Sec.~\ref{sec:GrayMaps}), blurs the output image by detecting too many edges.
From these results, we can tell the ONN-HAM algorithm correctly detects black and white edges, despite the double detection, while it is not enough selective on gray scale edges. Note, the double detection could be corrected with an additional post-processing algorithm.
 
\subsection{512x512 Standard Images}
\label{sec:Res512x512Mat}

To further validate our solution on black and white images, we also perform edge detection on 512x512 standard test images. We test with grayscale images called "cameraman," "lena," and "living room" converted into black and white.
Simulation of large scale images with the fully digital ONN design are not feasible due to Vivado simulation time and the sequential scanning. 
Scanning sequentially a 512x512 image with a 3x3 filter and a step of 1 pixel is equivalent to processing $260 100$ 3x3 images.
Thus, to be able to simulate large scale images, we simulate all possible combinations of black and white 3x3 images ($2^9 = 512$ possibilities) and use the corresponding output while scanning the large scale image on Matlab. Note, performing this test with 9-stage 3x3 gray scale images remains impossible, as there are $9^9 = 387 420 489$ possible inputs.

\begin{figure*}
    \centering
    \includegraphics[width=\linewidth]{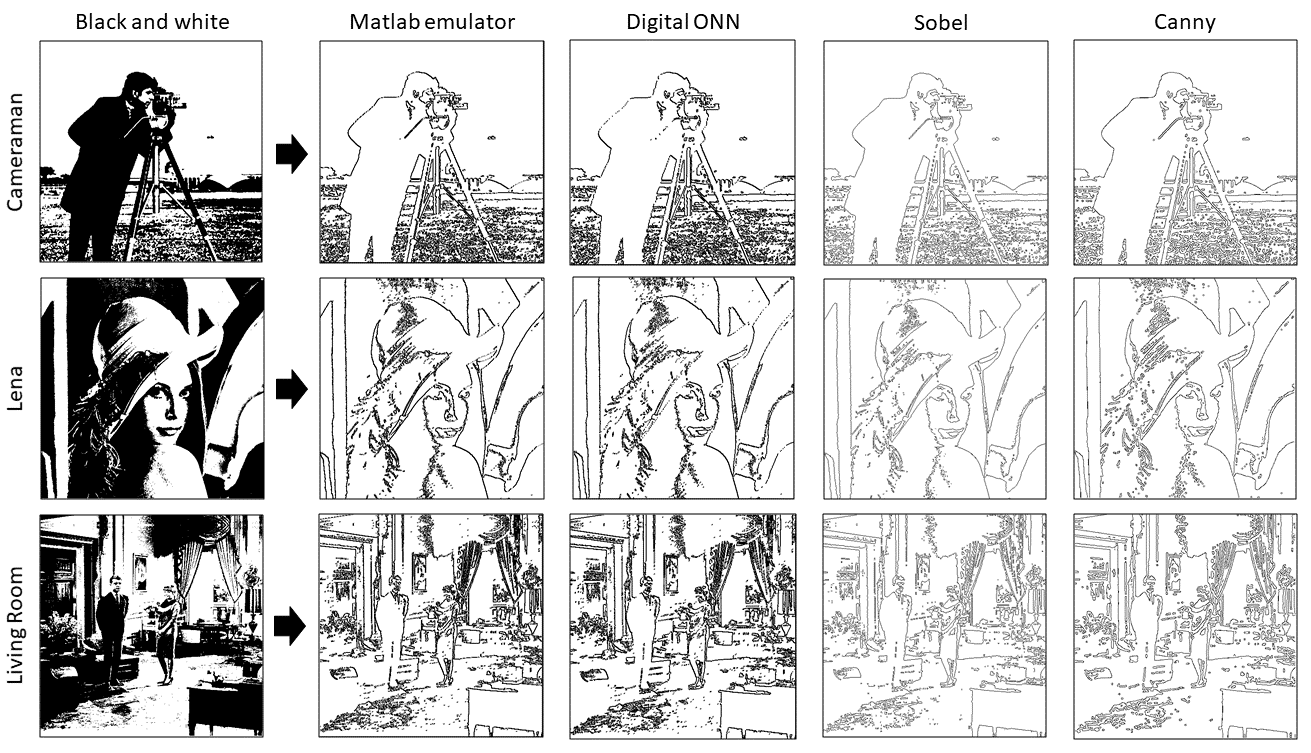}
    \caption{Comparison of output images obtained with the ONN-HAM solutions and standard Sobel and Canny algorithms performing edge detection on 512x512 black and white standard test images.}
    \label{fig:ONN-big-Comp}
\end{figure*}

Fig.~\ref{fig:ONN-big-Comp} presents output images obtained from ONN-HAM, Sobel and Canny edge detection algorithms. 
It validates the ONN-HAM efficiency to retrieve principal image edges. Note that some imperfections are due to the black and white conversion. 
Visually, we can state that our ONN-HAM solution has similar precision as Sobel and Canny algorithms on large-scale black and white images. Additional processing demonstrates that ONN-HAM makes double detection of edges as shown in previous tested images.

\section{Discussion}
\label{sec:Discussion}

Simulations processed with both ONN-HAM designs, the Matlab emulator and the fully digital ONN, highlighted the ability of our solution to detect edges. In this Section, we discuss limitations of our system and possible quality improvement. We also provide resource and timing characteristics of the fully digital design and evaluate its performances on real-time image processing. Finally, we state computation advantages of the ONN paradigm for this application. 

\subsection{Quality assessment}
\label{sec:QualityDisc}

Our ONN-HAM edge detection solution shows similar quality results as state-of-the-art Sobel and Canny algorithms on black and white images. 
Almost all edges, apart from left-oriented diagonal, are correctly detected. 
Note, a possible solution to counter the lack of left-oriented edge detection, would be to train the model with an additional training pattern.
However, we notice that our solution detects edges twice. 
This limits the edge detection efficiency on gray scale images but can also increase edge detection for uncleared edges. 
This effect can also impact the edge detection ability on noisy images. 
We believe that the behavior on noisy images requires further investigation as it is a critical point for edge detection algorithm, hence the additional gaussian filter in Canny algorithm. 
An additional post-processing pooling function could help to counter this double detection effect.

\subsection{Timing performances}
\label{sec:TimeDisc}

We extract timing and resource characteristics from simulations performed with the fully digital ONN design, and we evaluate real-time image processing feasibility. 
Table~\ref{tab:onnFPGAres} shows characteristics and performances extracted with the digital ONN design performing image edge detection. 
It indicates that each ONN-HAM with oscillators operating at $F_{osc}=2.7 MHz$ ($T_{osc}=6ns$ and $F_{sys} = 166 MHz$) on a 3x3 image needs $240 ns$ for initialization and $1 us$ to $2 us$ for stable computation. 
This means ONN converges to a memorized pattern in 2 to 5 oscillation cycles.
Note, unstable computation is detected after 10 oscillation cycles without stabilization. Stabilization is considered after two oscillation cycles with equal phase stages.

Furthermore, using a sequential scanning of the image, we estimate that our digital ONN-HAM solution on FPGA can treat 28x28 images in $1.5 ms$, respecting real-time cameras constraints of 25-30 images per second. 
Estimation is calculated using the ONN initialization and computation time multiplied by the number of 3x3 images to treat, as it is considered a sequential process. 
If we consider a camera with an output flow of 30 images per second, our solution can do real-time treatment of images of up to 120x120 pixel dimensions. 
Other reported FPGA implementations of Sobel or Canny algorithms allow faster image processing \cite{sobelFPGA1, cannyFPGA}, by performing highly parallel computation.

The main drawback of our ONN-HAM solution is its run time, but there is room for improving the ONN digital design \cite{OnnFPGA} for image processing timing requirements. 
First, we can increase the FPGA frequency, and consequently, ONN oscillation frequency. 
The current FPGA device allows a maximum system frequency of $F_{sys} = 175MHz$, however other more powerful FPGAs could be used to increase the system frequency. 
Also, in Table~\ref{tab:onnFPGAres}, we highlight that the ONN-HAM digital design uses small FPGA resources, with less than 1\% of the available FPGA resources. 
This first permits its integration in larger architectures. 
And second, it proposes an alternative to the sequential process, by using multiple ONNs in parallel to accelerate the image filtering process. 
For example, using 20 parallel ONNs with the same system frequency of $F_{sys}=166 MHz$, our ONN-HAM can process 512x512 images in $30 ms$, respecting real-time constraints. 

Finally, the actual fully digital ONN developed in \cite{OnnFPGA} is a fully connected ONN designed for AAM tasks. We believe it can be further adapted for HAM tasks. 

\begin{table*}[t]
    \centering
    \caption{Results obtained from simulations with the digital ONN design performing edge detection}
    \begin{tabular}{l | l}
        \toprule[2pt]
        \multicolumn{2}{l}{Single ONN characteristics} \\
        \midrule[1pt]
        System frequency ($F_{sys}$) & 166 MHz \\
        ONN size & 11 neurons (9 in, 2 out) \\
        ONN frequency & 2.734 MHz ($=F_{sys} / (16*4)$) \\
        Initialization time (ns) & 240 \\
        Computation time (us) & min: 1 - max: 2 \\
        LUTs (\%) & 402 (0.76) \\
        Flip-Flops (\%) & 443 (0.42) \\
        \midrule[2pt]
        \multicolumn{2}{l}{Estimations of full image sequential processing } \\
        \multicolumn{2}{l}{with single ONN (computation time: 2 us)} \\
        \midrule[1pt]
        Image size & Image time processing \\ \hline
        28x28  & 1.5 ms\\
        50x50  & 5.2 ms \\
        120x120  & 31.9 ms \\
        128x128  & 35.6 ms\\
        512x512  & 582.6 ms \\ 
        \bottomrule[2pt]
    \end{tabular}
    
    \label{tab:onnFPGAres}
\end{table*}

\subsection{Computation advantages}
\label{sec:CompDisc}

Despite the timing limitation, our ONN-HAM solution for image edge detection presents computation advantages. Only one ONN-HAM filter is necessary to detect multiple types of edges instead of using multiple convolutional kernels (and gaussian pre-processing for Canny algorithm). This limits the number of parameters from 18 for two convolutional kernels to 9 for our ONN-HAM solution. Consequently, decreasing the number of parameters reduces the memory space needed for implementation. 

Finally, ONN analog computing paradigm aims to provide a solution for low-power computation.
We believe, measuring power consumption using our fully digital ONN-HAM solution is not relevant because the ONN energy efficiency comes from its analog implementation. 
This work aims to show a proof of concept of a new solution to use ONN as HAM, which is efficient for edge detection application. 
Further investigation and experimentation are needed to report on ONN-HAM analog computation and energy efficiency.
However, research around low power analog ONN from novel materials, to novel devices, and up to circuit architectures encourages new system-level architectures and applications exploration \cite{lowpower, ONN1, ONN2, ONN4, ONNMap, ONNVO2}.

\section{Conclusion}
\label{sec:Conclusion}

In this paper, we proposed a solution to process hetero associative memory tasks using the ONN computing paradigm for the first time. We simulate ONN-HAM and apply it to perform image edge detection. We test and validate our system with two different designs, a software Matlab emulator based on HNN, and a hardware fully digital ONN design. 
We verify our edge detection solution first on gray scale evaluation maps, made of a white background with gray scale squares. Then, we validate it on more complex and realistic 28x28 gray scale and black and white MNIST images and finally, we confirm its efficiency on large scale 512x512 black and white standard test images. 

Our ONN-HAM solution is able to retrieve principal edges with only one filter and two training patterns on black and white as well as gray scale images. 
Despite a double edge detection phenomenon, we report similar results with our ONN-HAM solution to the standard classical Sobel and Canny algorithms, for black and white images. 
The double detection phenomenon appears due to no processing used to select stronger edges, as in Sobel and Canny algorithm.

We also display edge detection on gray scale images. 
Our ONN-HAM solution is highly sensitive to gray scale edges which is an advantage for soft edges, but limits the efficiency on realistic gray scale images where too many edges are detected.

Using ONN-HAM digital design running at a system frequency of $F_{sys}=166 MHz$, with an oscillation frequency of $F_{osc}=2.7 MHz$ $(T_{osc} = 6 ns)$, our system needs $240 ns$ for initialization of the 11 neurons (9 input neurons and 2 output neurons), and from $1 us$ to $2 us$ for computation, corresponding to 2 to 5 oscillation cycles. 
Considering sequential scanning of the image, we estimate the processing time of a 120x120 image to $31.9 ms$, respecting real time camera requirements of 30 images per second.

\section*{Acknowledgement}
This work was supported by the European Union’s Horizon 2020 research and innovation program, EU H2020 NEURONN project under Grant 871501 (www.neuronn.eu).

\bibliographystyle{abbrv}
\bibliography{mybib}

\end{document}